# CLASSIFICATION OF FUSED FACE IMAGES USING MULTILAYER PERCEPTRON NEURAL NETWORK

Debotosh Bhattacharjee[2], Mrinal Kanti Bhowmik[1], Mita Nasipuri[2], Dipak Kumar Basu[2], Mahantapas Kundu[2]
[1]Department of Computer Science and Engineering, Tripura University, Suryamaninagar- 799130, Tripura, India
[2]Department of Computer Science and Engineering, Jadavpur University,
Kolkata- 700032, India

*Abstract-* This paper presents a concept of image pixel fusion of visual and thermal faces, which can significantly improve the overall performance of a face recognition system. Several factors affect face recognition performance including pose variations, facial expression changes, occlusions, and most importantly illumination changes. So, image pixel fusion of thermal and visual images is a solution to overcome the drawbacks present in the individual thermal and visual face images. Fused images are projected into eigenspace and finally classified using a multi-layer perceptron. In the experiments we have used Object Tracking and Classification Beyond Visible Spectrum (OTCBVS) database benchmark thermal and visual face images. Experimental results show that the proposed approach significantly improves the verification and identification performance and the success rate is 95.07%. The main objective of employing fusion is to produce a fused image that provides the most detailed and reliable information. Fusion of multiple images together produces a more efficient representation of the image.

## I. INTRODUCTION

Human face recognition is a challenging task and its domain of application is very vast, covering different areas like security systems, defense applications, and intelligent machines. It involves different image processing issues like face detection, recognition and feature extraction [1] [2] [3]. Recently, fusion of face images is gaining acceptance as a superior biometric in case of face recognition. Data fusion is a method which combines different types of data gathered by the simultaneous use of several sensing modalities to generate a new type of data. Various perceptual mechanisms integrate these senses to produce the internal representation of the sensed environment. The integration tends to be synergistic in the scene that information inferred from the process cannot be obtained from any proper subset of the sense modalities. This property of synergism is one that should be sought for when implementing multisensor integration for machine perception. The principal motivation for the fusion approach is to exploit such synergism in the technique for combined interpretation of images obtained from multiple sensors.

So far, research work on fusion has been carried out for years. And the obtained fusion methods can be classified into two categories as stated in [11]. One is about weakly coupled fusion methods, and the other is about strongly coupled fusion methods. In the first category of fusion methods, fusion of data produced by sensory modules does not affect the operation of the modules. On the contrary, for strongly coupled fusion methods, the modules producing the data to be fused are being affected in some way by other information from other modules.

For the detailed review on current advances in visual and thermal face recognition refer to [12]. Simple image fusion in spatial domain is discussed in [13], where face recognition is used for testing the fusion of face database images. Both image fusion and decision fusion is employed in [14] to improve the accuracy of the face recognition system.

Recently, researchers have investigated the use of fusion of thermal infrared and visual face images for person identification to tackle the drawbacks of individual thermal and visual images [7] [8] [9] [10] [11]. In this paper, we present a novel approach to the problem of face recognition that realizes the full potential of the fusion of thermal IR band and visual band images. In this work at first thermal and visual face images are combined together and create the fused image of corresponding thermal and visual face images, after that using these transformed fused images eigenfaces are computed and finally those eigenfaces thus found are classified using a multilayer perceptron.

The organization of the rest of this paper is as follows. In section II, the overview of the system is discussed, in section III experimental results and discussions are given. Finally, section IV concludes this work.

## II. THE SYSTEM OVERVIEW

Here we present a technique for human face recognition. In this work we have used Object Tracking and Classification Beyond Visible Spectrum (OTCBVS) database benchmark thermal and visual face images. Every thermal face image and visual face image is first combined and converted into fused image. These transformed images are separated into two groups namely training set and testing set. The eigenspace is computed using training images. All the training and testing images are projected into the created eigenspace and named as fused eigenfaces. Once these conversions are done the next task is to use a classifier to classify them. A multilayer perceptron is used for this purpose. The block diagram of the system is given in figure 1. In this figure dotted line indicates feedback from different steps to their previous steps to improve the efficiency of the system e.g. if the classification results are not satisfactory some adjustment in the internal parameters may be done. In case of eigenspace projection number of eigenvectors, used to create eigenspace, may be increased or decreased to improve better representation of eigenfaces in the reduced space in order to achieve higher accuracy in recognition of faces.

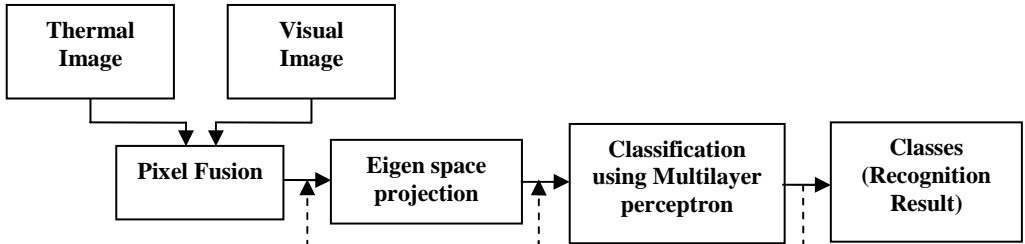

Figure 1: Block diagram of the system presented here.

### A. Thermal Infrared Face Images

Thermal infrared face images are formed as a map of the major blood vessels present in the face. Therefore, a face recognition system designed based on thermal infrared face images cannot be evaded or fooled by forgery, or disguise, as can occur using the visible spectrum for facial recognition. Compared to visual face-recognition systems this recognition system will be less vulnerable to varying conditions, such as head angle, expression, or lighting.

### B. Image Fusion Technique

The task of interpreting images, either visual images alone or thermal images alone, is an unconstraint problem. The thermal image can at best yield estimates of surface temperature that in general, is not specific in distinguishing between object classes. The features extracted from visual intensity images also lack the specificity required for uniquely determining the identity of the imaged object. The interpretation of each type of image thus leads to ambiguous inferences about the nature of the objects in the scene. The use of thermal data gathered by an infrared camera, along with the visual image, is seen as a way of resolving some of these ambiguities. In the other hand, thermal images are obtained by sensing radiation in the infrared spectrum. The radiation sensed is either emitted by an object at a non-zero absolute temperature, or reflected by it. The mechanisms that produce thermal images are different from those that produce visual images. Thermal image produced by an object's surface can be interpreted to identify these mechanisms. Thus thermal images can provide information about the object being imaged which is not available from a visual image [11].

A great deal of effort has been expended on automated scene analysis using visual images, and some work has been done in recognizing objects in a scene using infrared images. However, there has been little effort on interpreting thermal images of outdoor scenes based on a study of the mechanism that gives rise to the differences in the thermal behavior of object surfaces in the scene. Also, nor has been any effort been made to integrate information extracted from the two modalities of imaging.

In our method the process of image fusion is where pixel data of 70% of visual image and 30% of thermal image of same class or same image is brought together into a common operating image or now commonly referred to as a Common Relevant Operating Picture (CROP) [10]. This implies an additional degree of filtering and intelligence applied to the pixel streams to present pertinent information to the user. So image pixel fusion has the capacity to enable seamless working in a heterogeneous work environment with more complex data. For accurate and effective face recognition we require more informative images. Image by one source (i.e. thermal) may lack some information which might be available in images by other source (i.e. visual). So if it becomes possible to combine the features of both the images viz. visual and thermal face images then efficient, robust, and accurate face recognition can be developed.

We describe below in detail the fusion scheme considered in this work. We assume that each face is represented by a pair of images, one in the IR spectrum and one in the visible spectrum. Both images have been combined prior to fusion to ensure similar ranges of values.

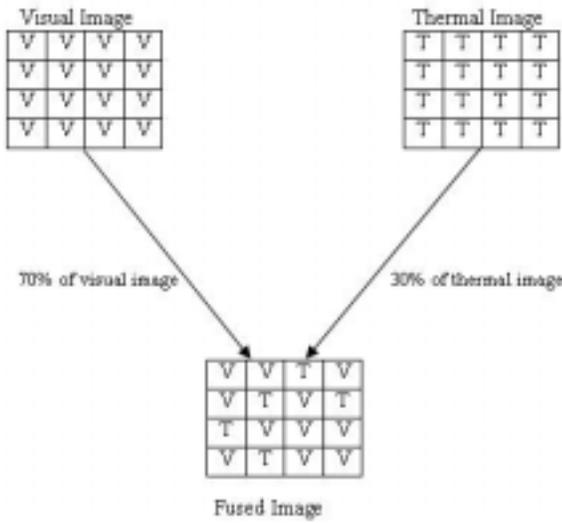

Figure 2: Fusion Technique.

We fused visual and thermal images. Ideally, the fusion of common pixels can be done by pixel-wise weighted summation of visual and thermal images.

$$F(x, y) = a(x, y)V(x, y) + b(x, y)T(x, y) \quad [12] \ldots\ldots\ldots\ldots\ldots\ldots\ldots\ldots\ldots\ldots(1)$$

where $F(x, y)$ is a fused output of a visual image, $V(x, y)$, and a thermal image, $T(x, y)$,
while $a(x, y)$ and $b(x, y)$ represent the weighting factors for visual and thermal images respectively. In this work, we have considered $a(x, y) = 0.70$ and $b(x, y) = 0.30$

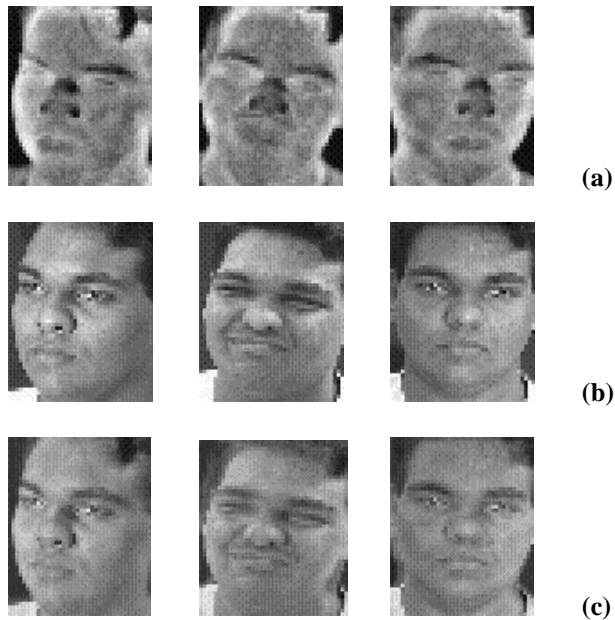

Figure 3: (a) Thermal Images, (b) Visual Images, (c) Fused Images of corresponding thermal and visual images.

C. **Eigenfaces For Recognition**

In mathematical terms, we wish to find principal components [4] [5] [6] of the distribution of faces, or the eigenvectors of the covariance matrix of the set of face images. These eigenvectors can be thought of as set of features which together characterize the variations between face images. Each image location contributes more or less to each eigenvector, so that we can display the eigenvector as sort of ghostly face which we call an eigenface. Each face image in the training set can be presented exactly in terms of a linear combination of the eigenfaces. The number of a possible eigenfaces is equal to the number of face images in the training set. However the faces can also be approximated using only the "best" eigenfaces-those that have the largest eigenvalues, and which therefore account for the most variance within the set face images. The best U eigenfaces constitute a U-dimensional subspace, which may be called as "face space" of all possible images. Identifying images through eigenspace projection takes three basic steps. First the eigenspace must be created using training images. After that all those training images are projected into the eigenspace and call them eigenfaces. Train a classifier using these eigenfaces. Finally, the test images are identified by projecting them into the eigenspace and classifying them by the trained classifier.

### D. ANN Using Backpropagation With Momentum

Neural networks, with their remarkable ability to derive meaning from complicated or imprecise data, can be used to extract patterns and detect trends that are too complex to be noticed by either humans or other computer techniques. A trained neural network can be thought of as an "expert" in the category of information it has been given to analyze. The Back propagation learning algorithm is one of the most historical developments in Neural Networks. It has reawakened the scientific and engineering community to the modeling and processing of many quantitative phenomena using neural networks. This learning algorithm is applied to multilayer feed forward networks consisting of processing elements with continuous differentiable activation functions. Such networks associated with the back propagation learning algorithm are also called back propagation networks.

### III: EXPERIMENTAL RESULTS AND DISCUSSIONS

This work has been simulated using MATLAB 7. For comparison of results experiments are conducted for fused images. A thorough system performance investigation, which covers all conditions of human face recognition, has been conducted. They are face recognition under i) variations in size, ii) variations in lighting conditions, iii) variations in facial expressions, iv) variations in pose.
We first analyze the performance of our algorithm using OTCBVS database which is a standard benchmark thermal and visual face images for face recognition technologies.

### A. OTCBVS Database

Our experiments were perform on the face database which is Object Tracking and Classification Beyond Visible spectrum (OTCBVS) benchmark database contains a set of thermal and visual face images. There are 700 images of visual and 700 thermal images of 16 different persons. For some subject, the images were taken at different times which contain quite a high degree of variability in lighting, facial expression *(open /* closed eyes, smiling /non smiling etc.), pose (Up right, frontal position etc.) and facial details (Glasses/ no Glasses). All the images were taken against a dark homogeneous background with the subjects in and upright, fontal position, with tolerance for some tilting and rotation of up to 20 degree. The variation in scale is up to about 10% all the images in the database.

### B. Classification Of Fused Eigenfaces Using Multilayer Perceptron

Out of total 700 thermal and visual images 400 images are taken out of which 200 are thermal images and 200 are visual images. Combining these thermal and visual images we get 200 fused images. 100 of these images are used as training set and rest 100 images are taken as testing images. The training set contains 10 classes which mean that each class has 10 images. Now 5 images from one particular class (which are not used as a training image) and 5 more images of the other classes are taken from the testing set. According to this process for all the 10 classes we get the result which is shown in the following graph of Figure 4. In this work a multilayer neural network with back propagation has been used. The learning algorithm error back propagation with momentum is used here.

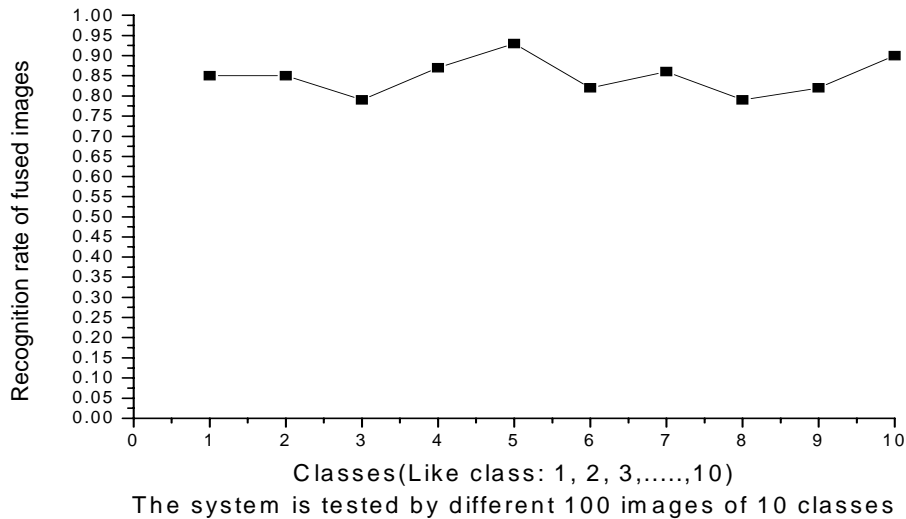

Figure 4: Study of recognition rate of fused images.

In order to access the effectiveness of the fused images we compare the fused face image recognition. Results obtained after applying the same procedure for fused image of thermal and visual images are shown in figure 4. Here, experiments are conducted for different number of images during testing.

**Table 1: Comparison of different fusion algorithms.**

| Fusion Scheme | Reference | Combining | Weights [IITV/IR (Image Intensified TV Sensor with Infrared Images)] |
|---|---|---|---|
| ADD (Adaptive Fusion) | [15] | None | 70%/30% |
| DWT1 (Discrete Wavelet Transform) | [21,16] | Weighted Average | 70%/30% |
| LAP1 (Laplacian Pyramid) | [17,16] | Weighted Average | 70%/30% |
| FSD1 (Filter Subtract Decimate) | [18,16] | Weighted Average | 70%/30% |
| GRAD1 (Gradient Pyramid) | [19,16] | Weighted Average | 70%/30% |
| MORPH1 (Morphological Pyramid) | [20,16] | Weighted Average | 70%/30% |
| SiDWT1 (Shift Invariant Discrete Wavelet Transform) | [22,16] | Weighted Average | 70%/30% |
| DWT2 (Discrete Wavelet Transform) | [21,16] | Choosing Maximum | None |
| LAP2 (Laplacian Pyramid) | [17,16] | Choosing Maximum | None |
| FSD2 (Filter Subtract Decimate) | [18,16] | Choosing Maximum | None |
| GRAD2 (Gradient Pyramid) | [19,16] | Choosing Maximum | None |
| MORPH2 (Morphological Pyramid) | [20,16] | Choosing Maximum | None |
| SiDWT2 (Shift Invariant Discrete Wavelet Transform) | [22,16] | Choosing Maximum | None |
| Present Method | Paper | Weighted Average | 30%/70% |

These algorithms (in Table 1) were tested for fusing low-light images produced by an image-intensified TV (IITV) sensor with infrared images produced by a Forward Looking Infrared (FLIR) camera, after registering these images as well as possible. Based upon the preliminary survey, we determined the most promising fusion approaches to be those given in Table 1. It is noted that the alternative of the combining method listed in Table 1 is employed at the lowest frequency band for the pyramid transform-based fusion, and low-low band for DWT and SiDWT fusion. The combining method for other frequency bands is "choosing maximum" in our tests. In the process of determining the methods in Table 1, we eliminated many algorithms. Some of them performed very poorly. The alternative of either the "choosing maximum" or the "weighted average" combining method at the lowest or low-low frequency band generates different MDB [17] fusion results with respect to contrast and brightness.

## IV. CONCLUSION

In this paper we have presented a human face recognition technique using fusion of pixel level fused image of thermal face and visual face results with varying illumination, facial expression, pose, and facial details. After the fusion of images as weighted sum, the fused images are projected into eigenspace. Those fused eigenfaces are classified using multilayer perceptron. Eigenspace is constituted by the images belong to the training set of the MLP.

The efficiency of our scheme has been demonstrated on Object Tracking and Classification Beyond Visible spectrum (OTCBVS) benchmark database and recognition rate obtained is 95.07%.

## V. ACKNOWLEDGMENT